\documentclass[sigconf, anonymous=false]{acmart}

\usepackage{subcaption} 
\usepackage{multirow} 
\usepackage{makecell} 
\usepackage{amsmath} 
\usepackage{balance} 

\AtBeginDocument{%
  \providecommand\BibTeX{{%
    \normalfont B\kern-0.5em{\scshape i\kern-0.25em b}\kern-0.8em\TeX}}}

\setcopyright{acmcopyright}
\copyrightyear{2020}
\acmYear{2020}
\acmDOI{}

\acmConference[BuildSys '20]{BuildSys '20: 7th ACM International Conference on Systems for Energy-Efficient Built Environments}{November 16--19, 2020}{Yokohama, Japan}
\acmBooktitle{BuildSys '20: 7th ACM International Conference on Systems for Energy-Efficient Built Environments,
  November 16--19, 2020, Yokohama, Japan}
\acmPrice{15.00} 
\acmISBN{978-1-4503-9999-9/18/06}

\begin{document}

\title{Balancing thermal comfort datasets: We GAN, but should we?}

\author{Matias Quintana}
\affiliation{%
  \department{School of Design and Environment}
  \institution{National University of Singapore}
  }
\email{matias@u.nus.edu}

\author{Stefano Schiavon}
\affiliation{%
 \department{Center for the Built Environment}
  \institution{University of California, Berkeley}
}
\email{schiavon@berkeley.edu}

\author{Kwok Wai Tham}
\affiliation{%
 \department{School of Design and Environment}
  \institution{National University of Singapore}
}
\email{bdgtkw@nus.edu.sg}

\author{Clayton Miller}
\affiliation{%
  \department{School of Design and Environment}
  \institution{National University of Singapore}
}
\email{clayton@nus.edu.sg}

\renewcommand{\shortauthors}{Quintana, et al.}

\begin{abstract}
Thermal comfort assessment for the built environment has become more available to analysts and researchers due to the proliferation of sensors and subjective feedback methods. These data can be used for modeling comfort behavior to support design and operations towards energy efficiency and well-being. By nature, occupant subjective feedback is imbalanced as indoor conditions are designed for comfort, and responses indicating otherwise are less common. This situation creates a scenario for the machine learning workflow where class balancing as a pre-processing step might be valuable for developing predictive thermal comfort classification models with high-performance. This paper investigates the various thermal comfort dataset class balancing techniques from the literature and proposes a modified conditional Generative Adversarial Network (GAN), \texttt{comfortGAN}, to address this imbalance scenario. These approaches are applied to three publicly available datasets, ranging from 30 and 67 participants to a global collection of thermal comfort datasets, with 1,474; 2,067; and 66,397 data points, respectively. This work finds that a classification model trained on a balanced dataset, comprised of real and generated samples from \texttt{comfortGAN}, has higher performance (increase between 4\% and 17\% in classification accuracy) than other augmentation methods tested. However, when classes representing discomfort are merged and reduced to three, better imbalanced performance is expected, and the additional increase in performance by \texttt{comfortGAN} shrinks to 1-2\%. These results illustrate that class balancing for thermal comfort modeling is beneficial using advanced techniques such as GANs, but its value is diminished in certain scenarios. A discussion is provided to assist potential users in determining which scenarios this process is useful and which method works best.
\end{abstract}



\maketitle

\section{Introduction}
\label{introduction}

Thermal comfort modeling has come a long way since one of its main predecessors, the Predicted Mean Vote (PMV) model \cite{Fanger1973}, which was proposed in the 1970s. Even though it is still the most used model in industry standards, current approaches are shifting to use direct feedback from the occupant, alongside relevant data, to figure out a model feasible of predicting the feedback for future or yet exposed scenarios \cite{Kim2018}. Thanks to the latest advancement in sensor and data management technologies, their reduction in cost, and the proliferation of interconnected devices known as the Internet of Things (IoT), building analysts and researchers are now able to collect more granular data, i.e., high sampling frequency, within buildings that can be used for this purpose as well as other building applications such as monitoring, control, and predictive maintenance \cite{Chourabi2012}. Current data-driven approaches aim to employ environmental \cite{Barrios2017, Gao2013}, physiological \cite{Liu2019}, and behavioral \cite{KIM201896} data as features for thermal comfort prediction and are predominant in recent thermal comfort modeling research.

Nevertheless, generalizability and high-performance are hard to achieve in these data-driven models, especially given the dataset's cohort and sample sizes regardless of the algorithm chosen \cite{Enescu2017}. These models rely on labeled data, i.e., data points where environmental and physiological parameters are paired up with the occupant's subjective thermal comfort feedback. Although the proliferation of IoT sensors in the built environment is vast \cite{Chourabi2012}, sensing occupants has its challenges. Particularly, getting the occupant to provide their subjective comfort feedback is a tedious task. Recent studies use mobile applications, computer programs, or emails \cite{tcs2019, Fay2017, Clear2018, Liu2019} to facilitate the occupant subjective feedback collection, but these methods are still limited by administrative overhead, cohort size, and users' behavior during the experiments, i.e., survey fatigue \cite{porter2004multiple}, which in turn could create concerns about how accurate their responses are \cite{Clear2018}. This results in datasets where people only report certain discomfort (e.g., ``warm'') and not other classes (e.g., ``too hot'', ``cool'', or ``cold''). One way to address this is to conduct targeted surveys \cite{targeted-surveyDuarte2020}, e.g., if not enough ``cold'' responses are available, a survey will be prompted to the user when the temperature surrounding it is below 15$^{\circ}$C or another arbitrary low temperature. This approach, however, is only applicable to future human studies. While thermal comfort applications such as building operation controls avoid the need of comfort feedback by prioritizing a temperature range deemed ``comfortable'' \cite{Enescu2017}, most occupant-centric approaches still require user data collection.

\subsection{Addressing imbalanced data}
Datasets with a disproportionate ratio of samples in each class are referred to as \textit{imbalanced} datasets. This issue is often ignored, or only a portion of the dataset is used \cite{Fay2017, Cheung2017}, which leads to a wasted effort in the data collection process. While the number of choices, i.e., the number of different class values, may contribute to a more uneven number of samples per class, a binary scenario would limit the thermal comfort dataset's versatility. This situation is particularly critical when we consider thermal comfort models' usability and outputs to design indoor environments better and control existing mechanisms for heating or cooling \cite{Park2018}. Simulation results in \cite{Jung2020} found that an occupant's thermal comfort characteristics are one of the most impactful parameters to determine the energy savings of comfort-driven control strategies. In these scenarios, the direction of discomfort, i.e., ``too cold'' or ``too hot'', is needed and can be directly captured when at least three classes (e.g., ``cool'', ``comfortable'', and ``hot'') are present in the thermal comfort feedback. 

One potential solution for imbalanced data is the use of generative models, especially Generative Adversarial Networks (GANs). GANs models have been widely deployed for many generative tasks using image datasets, and the results have been at par with state of the art. While some applications of GANs in the built environment have shown their potential \cite{Stinner2019, Zhao2019, Yan2020}, efforts on numeric tabular datasets are mostly done in datasets with binary classes on the dependant variable and failed to scale with multi-class variables \cite{Mariani2018}. Their application in the field of thermal comfort datasets is limited, and there has not been a thorough investigation of which class balancing technique works best for this application. This paper aims to explore the application of GANs in this context compared to the methods from previous work and ascertain in which situations balancing provides more or less value for the accuracy of the workflow.

\section{Related work and novelty}
\label{related-work}

In order to deal with the imbalanced datasets, researchers make use of a part of the modeling workflow toolkit called data augmentation. This strategy is a way of pre-processing the data to increase its size and diversity; however, it also offers the additional benefit of enabling models, trained on said pre-processed data, to be more robust to transformations of the model's input \cite{Nikolenko2019}. As mentioned, thermal comfort researchers sometimes discard data points with the predominant class value, known as undersampling \cite{Fay2017, Cheung2017}, or they generate synthetic data points based on the real dataset to balance the number of samples per class, a way of augmenting the data also known as oversampling \cite{Lipczynska2018, Liu2019}. The latter approach often relies on standard algorithms such as Synthetic Minority Over-sampling Technique (SMOTE) \cite{smote} and Adaptive Synthetic sampling approach (ADASYN) \cite{adasyn}. Nevertheless, other fields familiarized with imbalanced datasets, such as computer vision, have shown the potential to utilize other algorithms for generating synthetic data points known as Generative Models. This family of algorithms models the real data distribution $P_r$, by learning a distribution $P_\theta$, parameterized by $\theta$, that approximates it. Two approaches to achieve this would be to directly learn $P_\theta$ or to learn a function $g_\theta$ which transform a common distribution $Z$ (i.e., Uniform or Gaussian) such that $P_\theta \approx g_\theta(Z)$. Since these models aim to capture the data distribution, after being trained on a dataset, they are capable of \textit{generating} data points that were not found originally in the dataset but are likely to come from the same distribution. One type of model from the latter approach are Generative Adversarial Networks (GANs) \cite{Goodfellow2014}.

GANs can be thought of as a game on which two models, most of the time neural networks, compete against each other. The Generator \textit{G} uses a random noise vector as an input and tries to generate an instance that looks like it was drawn from the same distribution as the data. On the other hand, the Discriminator \textit{D} aims to distinguish between true samples from the data and samples generated by \textit{G}. Variations of this initial setup have allowed researchers to tackled various image-based problems like human pose generation \cite{Ma2017}, image quality enhancer \cite{Perarnau2016}, balancing image datasets \cite{Mariani2018}, just to name a few. Furthermore, when compared to other generative models, GANs have shown to produce more realistic samples than other well-known models, such as Adversarial Autoencoders \cite{Zhu2017}. Another popular and highly useful variant of GANs are \textit{conditional} GANs \cite{Mirza2014}. In this case, samples' generation and discrimination are conditioned on additional information, e.g., a class label. 

A survey done in \cite{Wang2019} encountered that applications outside computer vision have yielded promising results, but research related to the application of GANs in other areas is still somewhat limited. To our knowledge, GANs have not yet been used in the context of thermal comfort studies, and there are few applications of them in the built environment. Work done in \cite{Zhao2019} explores the usability of GANs on the imagery of urban areas to identify factors associated with cycling crashes in an effort to help urban planners design safer environments. Moreover, GANs have also been incorporated into the fault detection and diagnosis of chillers for augmenting such datasets \cite{Yan2020} since they are more suitable for extremely imbalanced datasets \cite{Douzas2018}. On the task of generating synthetic tabular data, current efforts deal with privacy concerns. Researchers aim to have synthetic, or generated, datasets that share the same characteristics as the original datasets and can therefore be used interchangeably for modeling purposes or shared with third parties. \cite{tableGAN-Park2018b, pateGAN-Jordon2019} show GANs able to generate synthetic data that preserves the descriptive statistics of the original data but enhances the anonymization of the original users. These approaches have gained popularity within the healthcare domain \cite{healthrecordsGAN-Che2017}. Another existing method is Tabular-GAN \cite{Xu2018a}, which focuses on the marginal distribution of the features using a more complex network architecture for the Generator \textit{G}. Their subsequent work CTGAN \cite{CTGAN-Xu2019} alleviates the model complexity of their approach, but the main focus remains to generate purely synthetic samples rather than correctly balancing the dataset. As mentioned earlier, the studies found that, in practice, GAN variants from image-based problems under-perform when the input data is a mixture of continuous and discrete features, which is the case for tabular data like thermal comfort studies. Thus, an active research trend is finding empirical and theory-backed GAN variants for non-image datasets.

To bridge the gap of generative methods for imbalanced and numerical thermal comfort datasets, and building on previous work 
\cite{GAN-Quintana2019}
, we propose comfortGAN, a conditional Wasserstein GAN with gradient penalty (cWGAN-GP) as a class balancing algorithm for data-driven thermal comfort modeling instead of commonly used methods. We assessed the performance of a balanced thermal comfort dataset, composed of generated and real samples, on a multi-class classification model, on scenarios where comfort feedback can take as much as seven distinct values, as well as a reduced version with only three possible values. This analysis's focus was not only to show improvements but also to determine which thermal comfort scenarios are improved more or less to guide the general application of class balancing in this context. We applied these techniques on three large open datasets that include comfort satisfaction, preference, and sensation objectives at different scales (3, 5, and 7-point scales). 

\section{Approach}
\label{approach}
\subsection{Datasets}
Three publicly available thermal comfort datasets were chosen, one laboratory-controlled experiment, and two field experiments. Among the latter experiments, one was collected by one research group in homogeneous conditions, and the other is an ensemble of many studies. The dataset \cite{Francis2019_dataset} (hereafter named controlled dataset) consists of a year-long three-hour session controlled experiment in Pittsburgh - Pennsylvania, U.S.A, where 77 participants' environmental and physiological measurements were taken alongside thermal comfort subjective feedback via a mobile application throughout a fixed temperature scheduled. The dataset used in \cite{Jayathissa2020} (hereafter named field dataset) consists of two weeks of intensive sampling of 30 participants in an educational building in Singapore. Occupants wore a smartwatch from which they gave a minimum of 100 subjective feedback responses, and the environmental measurements of the nearest fixed sensor in the building were paired to each response. Finally, the ASHRAE Global Thermal Comfort Database II (hereafter named Comfort Database) \cite{FoldvaryLicina2018} is a global joint effort to collect and harmonize numerous thermal comfort field studies systematically. Due to the treatment and data cleaning these studies went through, all of them can be considered one whole dataset. 

\subsection{Feature selection and pre-processing}
Data-driven thermal comfort models rely on a handful of measurable features and often outperform industry standards such as Predicted Mean Vote (PMV), especially when dealing with individual, personalized, thermal comfort. While the empirical models in the related literature can incorporate a handful of these variables, the exclusion of the rest can cause significant errors. Therefore, the features selected from each dataset will be the same as the features chosen in the related literature that used these datasets for data-driven thermal comfort modeling. By doing this, the data augmentation methods will evaluate how models would have changed if such augmentation models were to be used in the machine learning workflow. For the controlled dataset \cite{Francis2019_dataset}, we chose the feature set defined as \textit{Featureset-1} in \cite{tcs2019}, these features are \texttt{Air temperature}, \texttt{Skin temperature}, \texttt{Clothing insulation}, \texttt{Height}, \texttt{Shoulder circumference}, \texttt{Weight}, \texttt{Gender}, \texttt{Outside temperature}, \texttt{Outside relative humidity}, and \texttt{Thermal comfort feedback}.

For the field dataset \cite{Jayathissa2020} we chose their second set of proposed features with the same pre-processing cyclical representation of \texttt{Hour of the day} and \texttt{Day of the week}, the other features were \texttt{Heart rate}, \texttt{Relative humidity}, \texttt{Luminous intensity}, \texttt{Noise level}, \texttt{Air temperature}, and \texttt{Thermal sensation feedback}. 

Finally, for the Comfort Database \cite{FoldvaryLicina2018}, we picked six of the top seven most significant variables for data-driven thermal comfort proposed by \cite{Luo2020} for this dataset, which overlaps with the already chosen features from the other datasets. The features are \texttt{Standard Effective Temperature (SET)}, \texttt{Clothing level}, \texttt{Metabolic rate}, \texttt{Air temperature}, \texttt{Relative humidity}, \texttt{Air velocity}, and \texttt{Thermal sensation feedback}. We left out \texttt{Gender}, the extra feature \cite{Luo2020} chose because the available data points with this value were $\sim$40\% of the total data points in the database. The six features we chose resulted in $\sim$66,000 data points ($\sim$62\% of the entire database), and the \texttt{Thermal sensation feedback} was rounded to the closest integer. Overall, we believe the chosen features for all three datasets are a subset of common measurements researchers consider when modeling thermal comfort from longitudinal studies.

\begin{figure*}[!htp]
    \centering
    \begin{subfigure}{1\textwidth}
        \includegraphics[width=1\linewidth]{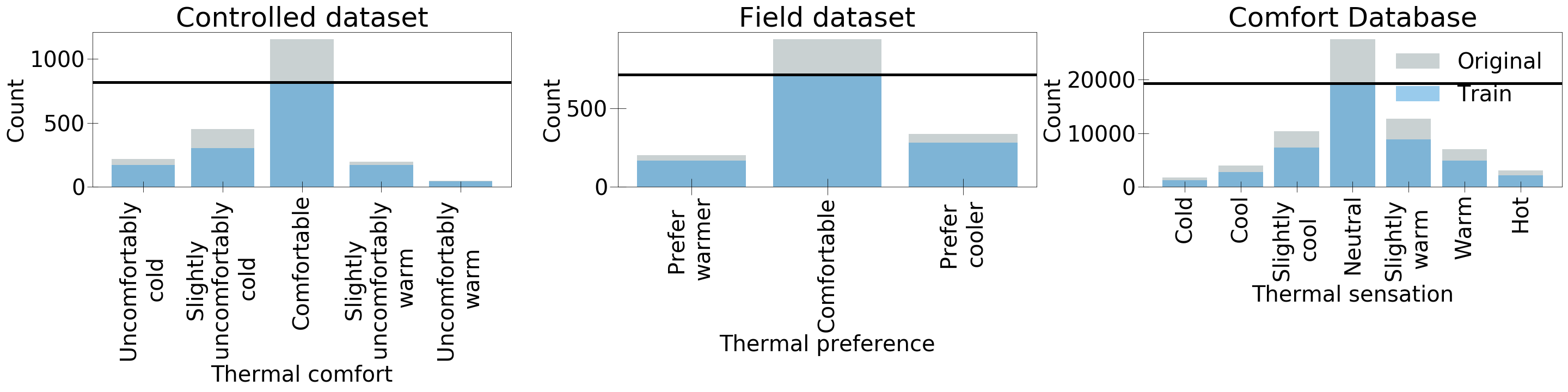}
        \caption{Original number of classes}
        \label{fig:datasets_dist_all_classes}
    \end{subfigure}
    \hspace{.35cm}
    \begin{subfigure}{1\textwidth}
        \includegraphics[width=1\linewidth]{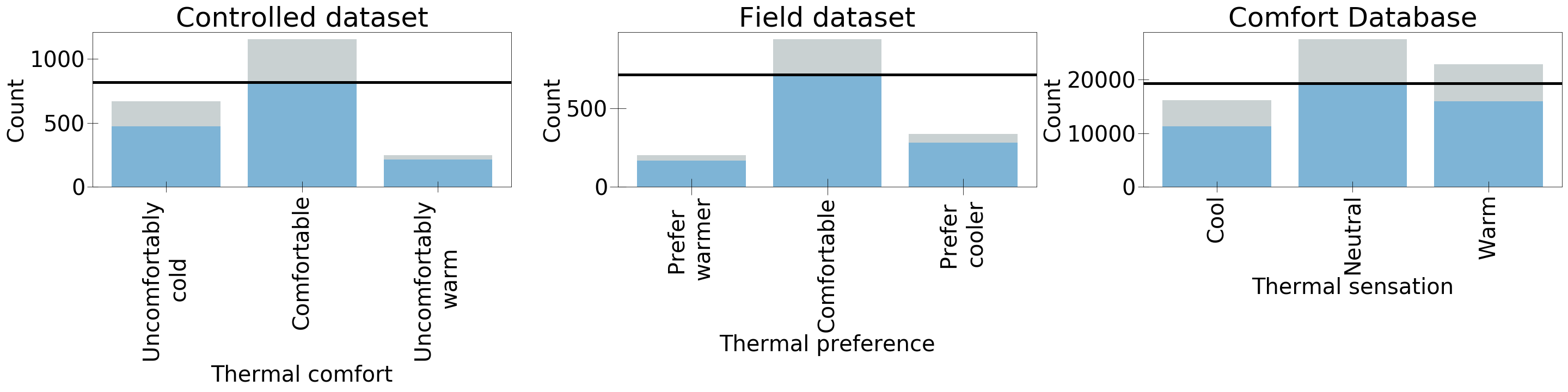}
        \caption{Classes remapped to only three values}
        \label{fig:datasets_dist_three_classes}
    \end{subfigure}
    \caption{Class distribution of original and train split set for the controlled dataset \cite{Francis2019_dataset}, field dataset \cite{Jayathissa2020}, and the Comfort Database \cite{FoldvaryLicina2018}, from left to right respectively. Row \textit{a)} shows the original datasets with their original classes and \textit{b)} is the datasets with their classes reduced to three. The original imbalanced dataset is colored grey and the train set, based on a 0.7 train-test random split, is colored blue. The horizontal black line indicates the number of data points of the predominant class in each dataset, and represent the value at which the other classes will be augmented.}
    \label{fig:datasets_dist}
\end{figure*}

\begin{figure}[htp]
    \centering
    \includegraphics[width=1\linewidth]{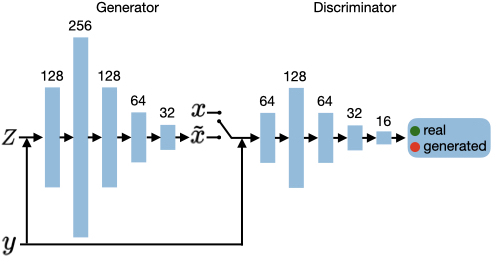}
    \caption{Overview of the architecture used. The Generator samples from a Normal distribution $Z$ and the class label $y$ is concatenated to the input of both the Generator and Discriminator (i.e., conditional GAN architecture variant). The latter takes real samples $x$ or generated samples $\tilde{x}$. Both the Generator and Discriminator are Multi-layer Perceptrons, where each blue rectangle is a layer, and the number on top indicates its number of units. Wasserstein loss with gradient penalty \cite{WGANimproved-Gulrajani2017} was chosen for more stable training.}
    \label{fig:architecture}
\end{figure}

\begin{figure}[!htp]
    \centering
    \includegraphics[width=1\linewidth]{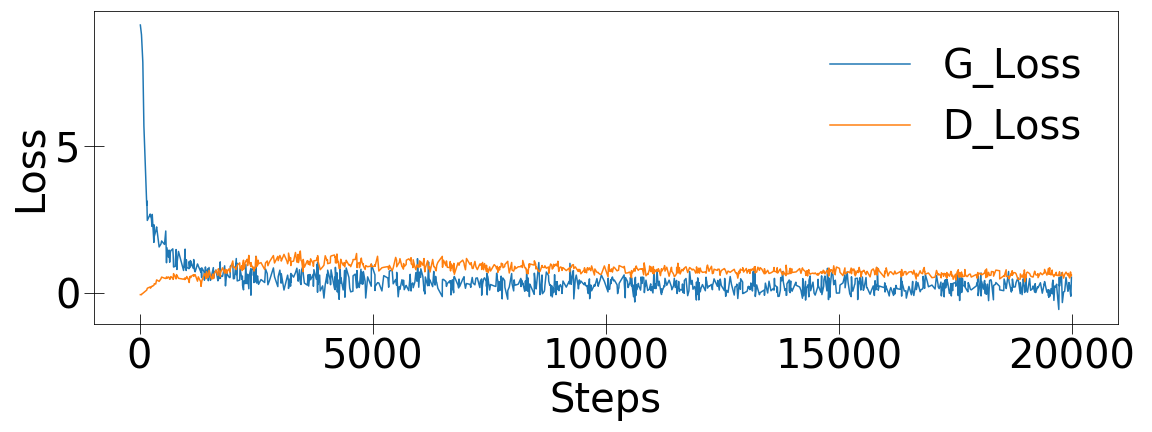}
    \caption{Loss value of both Generator \textit{(G\_Loss)} and Discriminator \textit{(D\_loss)} during the convergence process on the training split of the controlled experiment \cite{Francis2019_dataset}. Similar values were obtained for the other two datasets and can be found in the TensorBoard platform that accompanies this work.}
    \label{fig:convergence}
\end{figure}

\subsection{Customized GAN for thermal comfort}
\label{comfortGAN}
A survey done on GANs by \cite{Wang2019} identifies two main families of GANs: Architecture variant and loss variant. Architecture variants refer to the different types of neural networks used, e.g., fully connected networks, convolutional neural networks, recurrent neural networks, and the interaction between the data and the labels with the generator and discriminator. On the other hand, loss variants refer to the loss functions used to facilitate more stable learning of the Generator. For both families, the specific type and function to be used usually depends on the application \cite{Wang2019}. As introduced in Section \ref{related-work}, conditional GANs (cGAN) are an architecture variant where the generation and discrimination of samples is conditioned on the class label. By doing this, the Generator is exposed to a set of inputs for a specific class value or, in our case, a specific thermal comfort response. 

In term of the loss function, the original proposed loss \cite{Goodfellow2014} can lead to vanishing gradients, stopping the training process, unless the distributions of real and generated samples have significant overlap \cite{Wang2019}. However, in practice, the possibility that the real samples distribution and the generated samples distribution do not overlap or have negligible overlap is very high \cite{Arjovsky2019}. Therefore, a loss variant known as Wasserstein GAN (WGAN) \cite{WGAN-Arjovsky2017} was introduced to facilitate the training and convergence of both the Generator and Discriminator. Subsequent modifications on WGAN, known as WGAN-gradient penalty (WGAN-GP) \cite{WGANimproved-Gulrajani2017}, enhances training stability and have shown better results and convergence in practice compared to conventionally used image-based GAN variants (e.g., convolutional GANs), specifically on tabular data from other fields \cite{CTGAN-Xu2019, Yan2020}. Therefore, we move away from the vanilla architecture used in 
\cite{GAN-Quintana2019}
and on this work, we use the WGAN-GP loss variant for comfortGAN.

For categorical features as well as for the labels, we proceeded to smooth them by converting them to one-hot-encoding representation and adding uniform noise ($Uniform(0,\gamma)$, for which we chose $\gamma$=0.2) to each resulting binary variable and re-normalize each representation, similar to \cite{Xu2018a}. On the other hand, continuous features are scaled in the range of $(-1, 1)$. Recent research shows that the capacity and performance of GANs are related to the network size and batch size \cite{Brock2019}. With this in mind, we proceeded to evaluate different combinations of parameters on the datasets. Figure \ref{fig:datasets_dist} shows the distribution of class values for the original datasets and the distribution of their respective 0.7 train-test ratio train set, which is used for experiments throughout this paper. Figure \ref{fig:datasets_dist_all_classes} shows the original distribution with all possible class values whereas Figure \ref{fig:datasets_dist_three_classes} shows a reduced version on which the extremes (``cold'' or ``hot'') are grouped into one single class in order to have only 3 classes. This reduction only takes place in the controlled dataset \cite{Francis2019_dataset} and in the Comfort Database \cite{FoldvaryLicina2018}, which initially had 5 and 7 classes, respectively.

After trial-and-error, the architecture was set to five fully connected layers with \texttt{Rectified Linear Units (ReLU)} as activation functions for the Generator and \texttt{Leaky ReLU} with a slope of 0.2 for the Discriminator; the number of units per layer is 128/256/128/64/32 and 64/128/64/32/16 respectively, with a learning rate set to 0.0002, Figure \ref{fig:architecture} shows the architecture used. Following the suggestions in \cite{WGANimproved-Gulrajani2017}, the optimizer used for both the Generator and Discriminator was \texttt{Adam}, \texttt{Dropout} regularization of 0.5 was used on the Discriminator, and \texttt{Batch Normalization} was used on the Generator.

We tried different combinations of latent dimension size, batch size, and number of critics for the Discriminator (i.e., number of times the Discriminator is trained per Generator training iteration) for each training set. Ultimately, we chose the parameters that visually showed convergence after 20,000 (controlled and field dataset) and 200,000 (Comfort Database) iterations on the Generator. More iterations were used for the Comfort Database to account for the size of the dataset. These numbers were chosen empirically after revising the number of iterations in the related work \cite{CTGAN-Xu2019, Yan2020} ranges from 18,000 to 40,000 for medium-sized tabular datasets. While there is no explicit guarantee this range of iterations is suitable for all thermal comfort datasets, existing work in GANs for tabular data suggests these ranges are an excellent place to start while the theoretical explanations are a research trend.

The convergence plots of the Discriminator and Generator losses are available online via the TensorBoard platform\footnote{https://tensorboard.dev/experiment/heo6pFrNTpGNW4gVGe9Dhw/} and one of the plots available at TensorBoard is displayed in Figure \ref{fig:convergence}. This Figure shows the Generator and Discriminator losses, \texttt{G\_loss} and \texttt{D\_loss} respectively, for the controlled dataset \cite{Francis2019_dataset}. The final values of batch size, number of critics, and latent dimension size for the controlled dataset \cite{Francis2019_dataset}, field dataset \cite{Jayathissa2020}, and Comfort Database \cite{FoldvaryLicina2018} are 128, 1, and 20; 128, 3, and 80; 64, 1, and 100; respectively. Finally, while existing GAN architectures and loss functions were used, this work's technical contribution extends to the appropriate selection of them and evaluation of the framework for three large open public datasets in the context of thermal comfort.

\subsection{Evaluation}
\label{evaluation}
Evaluating generative models can be tricky since it is rarely a straightforward process, especially when different metrics can yield substantially different results \cite{Theis2015, Zhu2017}. If the goal is image generation, a subjective evaluation based on visual fidelity, e.g., ``real vs. fake'' perceptual study, can be appropriate and can even be crowd-sourced on Amazon Mechanical Turk (AMT) \cite{Zhu2017}. For the case of numeric datasets, it is common to rely on the effect that generated samples, combined with real samples, have on the performance of a classification model \cite{Mariani2018, tableGAN-Park2018b, Xu2018a, Yan2020}, researchers often name this \textit{machine learning efficacy}. Moreover, work done in \cite{Mariani2018} added metrics to evaluate the generated samples themselves, i.e., how diverse they are and if they look similar to the original training samples in an image dataset.

\begin{figure}[htp]
    \begin{subfigure}{1\columnwidth}
        \centering
        \includegraphics[width=0.55\textwidth]{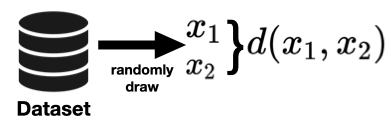}
        \caption{Variability of generated samples: Difference between generated samples.}
        \label{fig:evaluation_diagrams_var}
    \end{subfigure}%
    \vspace{0.25cm}
    \begin{subfigure}{1\columnwidth}
        \centering
        \includegraphics[width=0.65\textwidth]{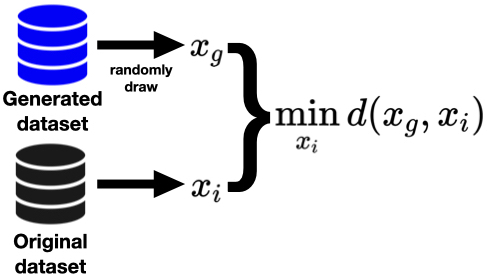}
        \caption{Diversity of generated samples with respect to the training set: Difference between closest pairs from generated and training set.}
        \label{fig:evaluation_diagrams_div}
    \end{subfigure}
    \vspace{0.25cm}
    \begin{subfigure}{1\columnwidth}
        \centering
        \includegraphics[width=1\textwidth]{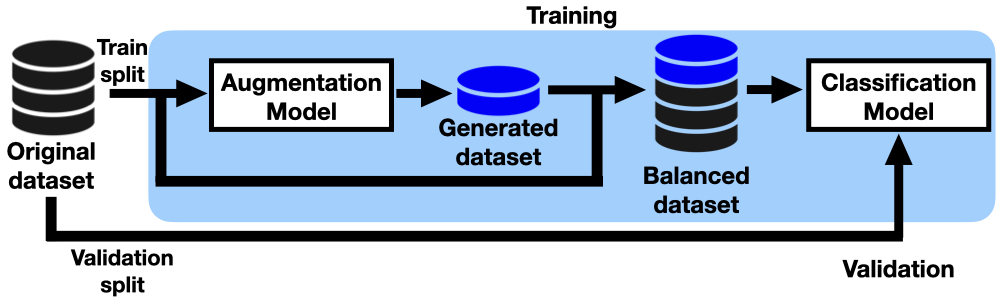}
        \caption{Machine learning efficacy or quality of generated samples on classification tasks: Classification accuracy on a balanced, mixture of generated and training samples, dataset.}
        \label{fig:evaluation_diagrams_ml}
    \end{subfigure}
    \caption{Diagrams for the different evaluation metrics used to evaluate the generated samples for all augmentation methods}
    \label{fig:evaluation_diagrams}
\end{figure}

We want to make sure our GAN variant does not suffer from a well known GAN-related problem called \textit{mode collapse}, i.e., generating the same type of sample over and over again. Thus, we quantify the \textit{variability of generated samples} by randomly drawing two generated samples and calculating their Euclidean distance to quantify how different the samples are. In image problems, this distance is usually calculated by measuring the Structural Similarity Index (SSIM) \cite{Mariani2018}. Secondly, to make sure the algorithm is not just memorizing the training samples, we measure the \textit{diversity of generated samples with respect to the training set} by randomly drawing a sample from the generated set and finding its closest matching sample, i.e., minimum Euclidean distance, from the training set. Finally, as our goal is to increase the minority classes on the original dataset to make classification models more robust, we combine the training set and the generated set to form a balanced dataset and train a classification model. We then report the classification accuracy, F1-micro score for this multi-class scenario; on the test set, we refer to this as the quality of generated samples on classification tasks, or \textit{machine learning efficacy}. Figure \ref{fig:evaluation_diagrams} shows a visual representation of the mentioned metrics. 

The classification model chosen for all the evaluation metrics is the Decision Random Forest (RDF) based on their extensive use on the thermal comfort modeling literature and in the respective work that utilizes the datasets \cite{tcs2019, Jayathissa2020}; they were also found to be the top-performing model for data-driven thermal comfort modeling on the Comfort Database \cite{Luo2020}. One RDF classifier was used per dataset, and their hyper-parameters were grid-searched based on 10-fold cross-validation on each respective train set. Experiments showed that 100 trees seem to consistently provide classifiers with the highest average cross-validation accuracy across all training sets, and this result is aligned with the models the authors used on the respective datasets. Finally, the optimal depth varied from 9 to 13, which coincide with what was found in \cite{Luo2020}. Thus, 100 trees and a tree depth of 10 were fixed for all classifiers; by specifying the classifier, we emphasize the training samples' role in the model performance. Furthermore, to imitate a conventional data-driven modeling approach, we used a random train-test split (Figure \ref{fig:datasets_dist}) and will evaluate the classification models' performance on said test split. Related work has opted to pick a test set carefully, comprise of real samples, with balanced classes \cite{Yan2020}; however, we see that in practice, thermal comfort researchers don't have the luxury of having enough samples per class to build a balanced test set due to experiment's cohort sizes or the number of data points per participants. Although the test set will also be inevitably imbalanced, we chose this approach to be as close as possible to what thermal comfort researchers do in practice.

Each draw action is repeated 30 times for each class value, and the average distance is reported on both scenarios, variability, and diversity. The baselines for all three metrics are calculated with the original train set alone, without any generated sample, which translates to the baseline of \textit{machine learning efficacy} being the performance of the fixed RDF model trained on the original imbalanced train set. Furthermore, a total of four other augmentation models are evaluated for comparison: SMOTE \cite{smote} and ADASYN \cite{adasyn} as models regularly used in the literature for data-driven thermal comfort modeling \cite{Lipczynska2018, Liu2019}, and TGAN \cite{Xu2018a} and CTGAN \cite{CTGAN-Xu2019} as GAN variant approaches for data generation. We repeated all evaluation metrics 30 times for each augmentation model; each repetition involved generating a newly generated dataset, and the average results are reported. The dimensions of each generated dataset are equal to the number of samples per class required to balanced the respective training set in Figure \ref{fig:datasets_dist}. We share our code base for reproducibility on a GitHub repository:
\begin{anonsuppress}
https://github.com/buds-lab/comfortGAN
\end{anonsuppress}

\section{Experimental Results}
\label{experiments}

\begin{table*}[!htb]
    \caption{Experiment results for all data augmentation models and datasets. The F1-micro score was used for classification evaluations (Machine learning efficacy), and Euclidean distance was used for distance calculation (Variability and Diversity). The datasets columns under each metric (first row in the column header) correspond to the controlled dataset \cite{Francis2019_dataset}, field dataset \cite{Jayathissa2020}, and the Comfort Database \cite{FoldvaryLicina2018}, respectively in that order. The percentage of improvement over the baseline is shown in parenthesis for the Machine learning efficacy case. For all metrics, the higher the value the \textbf{better}.}
    \centering

        
        \begin{tabular}{p{3cm} p{.8cm} p{.8cm} p{.8cm} p{.8cm} p{.8cm} p{.8cm} p{1.5cm} p{1.5cm} p{1.5cm}}
        \toprule
        \multirow{2}{*}{Model} & 
        \multicolumn{3}{c}{Variability} &
        \multicolumn{3}{c}{Diversity} &
        \multicolumn{3}{c}{Machine learning efficacy} \\
        \cmidrule(lr){2-4} \cmidrule(lr){5-7} \cmidrule(lr){8-10}
        & \cite{Francis2019_dataset} & \cite{Jayathissa2020} & \cite{FoldvaryLicina2018} &
        \cite{Francis2019_dataset} & \cite{Jayathissa2020} & \cite{FoldvaryLicina2018} &
        \cite{Francis2019_dataset} & \cite{Jayathissa2020} & \cite{FoldvaryLicina2018} \\
        \toprule
        Baseline                  & 52.50 & 175.32 & 18.12 &  1.90 &  5.66 & 0.55 & 0.60 & 0.65 & 0.26 \\
        SMOTE \cite{smote}        & 49.30 & 161.13 & 17.90 &  1.94 &  2.77 & 0.29 & 0.47(-13\%) & 0.53(-12\%) & 0.30(+4\%) \\
        ADASYN \cite{adasyn}      & 49.46 & 186.61 & 17.76 &  2.52 &  3.26 & 0.30 & 0.41(-19\%) & 0.51(-14\%) & 0.30(+4\%) \\
        TGAN \cite{Xu2018a}       & 42.80 & 142.26 & 17.99 & 15.58 & 12.10 & 0.89 & 0.62(+2\%) & 0.66(+1\%) & 0.38(+12\%) \\
        CTGAN \cite{CTGAN-Xu2019} & \textbf{62.85} & 159.27 & 17.40 & 23.92 & 19.80 & 1.17 & 0.48(-12\%) & 0.63(-2\%) & 0.40(+14\%) \\
        comfortGAN                & 56.67 & \textbf{324.07} & \textbf{38.95} & \textbf{34.21} & \textbf{21.27} & \textbf{17.26} & \textbf{0.64(+4\%)} & \textbf{0.67(+2\%)} & \textbf{0.43(+17\%)} \\
        \bottomrule
        \end{tabular}
        \subcaption{Original number of classes}
        \label{tab:experiments-results-all}
        \hspace{1cm}
        
        \begin{tabular}{p{3cm} p{.8cm} p{.8cm} p{.8cm} p{.8cm} p{.8cm} p{.8cm} p{1.5cm} p{1.5cm} p{1.5cm}}
        \toprule
        \multirow{2}{*}{Model} & 
        \multicolumn{3}{c}{Variability} &
        \multicolumn{3}{c}{Diversity} &
        \multicolumn{3}{c}{Machine learning efficacy} \\
        \cmidrule(lr){2-4} \cmidrule(lr){5-7} \cmidrule(lr){8-10}
        & \cite{Francis2019_dataset} & \cite{Jayathissa2020} & \cite{FoldvaryLicina2018} &
        \cite{Francis2019_dataset} & \cite{Jayathissa2020} & \cite{FoldvaryLicina2018} &
        \cite{Francis2019_dataset} & \cite{Jayathissa2020} & \cite{FoldvaryLicina2018} \\
        \toprule
        Baseline                  & \textbf{53.73} & 175.32 & 18.16 &  1.69 &  5.66 & 0.48 & 0.66 & 0.65 & 0.49 \\
        SMOTE \cite{smote}        & 50.19 & 161.13 & 18.28 &  1.18 &  2.77 & 0.20 & 0.60(-6\%) & 0.53(-12\%) & 0.50(+1\%) \\
        ADASYN \cite{adasyn}      & 50.52 & 186.61 & 17.12 &  2.13 &  3.26 & 0.20 & 0.53(-13\%) & 0.51(-14\%) & 0.50(+1\%) \\
        TGAN \cite{Xu2018a}       & 39.91 & 142.26 & 17.69 & 18.75 &  12.10 & 0.55 & 0.69(+3\%) & 0.66(+1\%) & 0.50(+1\%) \\
        CTGAN \cite{CTGAN-Xu2019} & 52.65 & 159.27 & 17.76 & 19.57 & 19.80 & 0.74 & 0.59(-7\%) & 0.63(-2\%) & 0.50(+1\%) \\
        comfortGAN & 43.84 & \textbf{324.07} & \textbf{37.07} & \textbf{37.75} & \textbf{21.27} & \textbf{13.13} & \textbf{0.72(+6\%)} & \textbf{0.67(+2\%)} & \textbf{0.51(+2\%)} \\
        \bottomrule
        \end{tabular}
        \subcaption{Classes remapped to only three values}
        \label{tab:experiments-results-reduced}
    \label{tab:experiments-results}
\end{table*}

\subsection{Experiment setup}
The workstation used for the experiments of this study has the following configuration: Intel Core (TM) i5-7600 @ 4.1GHz, NVIDIA GeForce GTX 1070 graphics card, Python 3.7.3 (64-bit), and PyTorch 1.5.0. The hyper-parameters tuning for TGAN \cite{Xu2018a} was done utilizing its random hyper-parameters search\footnote{https://sdv-dev.github.io/TGAN/readme.html\#model-parameters}, the best set of values over 30 random searches was chosen for each dataset; CTGAN \cite{CTGAN-Xu2019} does not provide such an environment for hyper-parameter tuning, so the default recommended hyper-parameters were chosen. As mentioned, the experiments were repeated with two copies of all datasets, one with the original number of classes (5, 3, and 7 for the controlled dataset \cite{Francis2019_dataset}, field dataset \cite{Jayathissa2020}, and Comfort Database \cite{FoldvaryLicina2018}, respectively) and another one with the classes remapped to only three values (the field dataset \cite{Jayathissa2020} remained unchanged).

\subsection{Variability of generated samples}
The baseline value on this first column \textit{Variability} on Table \ref{tab:experiments-results} gives a reference of how different, on average, the original samples are in the original training set. We want the generated samples to be as variable as the original samples used from training, although the main concern is to avoid a very low number. SMOTE \cite{smote} and ADASYN \cite{adasyn} score slightly below the baseline, with the exception of ADASYN \cite{adasyn} on the field dataset (middle column within the \textit{Variability} column). Both algorithms use a linear interpolation on the training data to generate new data points; thus, it is reasonable for them to score close to the baseline for both groups of datasets with original and reduced classes. 
The GAN variant models, CTGAN \cite{CTGAN-Xu2019} and TGAN \cite{Xu2018a} outperform the latter models with some exceptions on the Comfort Database \cite{FoldvaryLicina2018} for CTGAN \cite{CTGAN-Xu2019} and TGAN \cite{Xu2018a}. CTGAN \cite{CTGAN-Xu2019} achieves the highest score for the controlled dataset, in the original number of classes scenario and is the closest model to the baseline on the controlled dataset with reduced classes. ComfortGAN achieves the second-highest value for the comfort dataset with its original classes but drops to the last place when the classes are reduced. Nevertheless, it surpasses all the models for the other two datasets in both scenarios.

\subsection{Diversity of generated samples with respect to training set}
The second column \textit{Diversity} on Table \ref{tab:experiments-results} shows the average distance between a generated sample and the closest sample from the training dataset. The baseline in this metric is a reference on the average differences between similar samples, analogous as the previous metric; the main concern is to avoid a very low number. SMOTE \cite{smote} score close but below the baseline for almost all cases (except on controlled dataset \cite{Francis2019_dataset} with the original classes). Although ADASYN \cite{adasyn} also scores below the baseline for almost all cases (except on controlled dataset \cite{Francis2019_dataset}, it surpasses SMOTE \cite{smote} in all cases (with a tie-on Comfort Database \cite{FoldvaryLicina2018} with reduced classes). This result highlights the main difference among them; even if both rely on linear interpolation on the training samples, ADASYN \cite{adasyn} has a small random coefficient that allows it to produce samples that are slightly more different than the original samples in the training set. On the other hand, the GAN variants can achieve values one magnitude of order higher on all datasets with original and reduced classes. Although comfortGAN surpasses CTGAN \cite{CTGAN-Xu2019} and TGAN \cite{Xu2018a} on the controlled dataset (original and reduced classes), CTGAN \cite{CTGAN-Xu2019} and comfortGAN perform similarly on the field dataset \cite{Jayathissa2020}. On the Comfort Database \cite{FoldvaryLicina2018}, comfortGAN seems to develop more sparse data points, allowing for a higher distance, since visual inspection of the generated samples suggest that the numerical values are within the physical boundaries of the features. Nevertheless, the higher values of all GAN variants in terms of distance of generated versus original samples is expected given that the objective of these methods is to try to learn the underlying data distribution instead of only doing minor changes on the original training data points, like the traditional approaches like SMOTE \cite{smote} and ADASYN \cite{adasyn}.

\subsection{Machine learning efficacy}
Finally, we assess the accuracy of a classifier trained on the balanced dataset, i.e., a dataset with both the training samples and the generated ones such that all classes have the same number of data points. Given that the datasets used in this work are multi-class, the accuracy metric is calculated as an F1-micro score as it was also the metric reported related work on these datasets \cite{tcs2019, Jayathissa2020, Luo2020}. The last column \textit{Machine learning efficacy} in  Table \ref{tab:experiments-results} shows that generated samples from SMOTE \cite{smote} and ADASYN \cite{adasyn} end up affecting the classification performance on all datasets and only barely increasing the performance on the Comfort Database \cite{FoldvaryLicina2018} with reduced classes. When the datasets have more than three classes (controlled dataset \cite{Francis2019_dataset} and Comfort Database \cite{FoldvaryLicina2018}, Table \ref{tab:experiments-results-all}), comfortGAN increases the classifier performance 4\% and 17\% respectively. TGAN \cite{Xu2018a} increase the performance on the controlled dataset \cite{Francis2019_dataset} by 2\% and both TGAN and CTGAN \cite{CTGAN-Xu2019} increase the performance on Comfort Database \cite{FoldvaryLicina2018} by 12\% and 14\% respectively. However, when the number of classes is reduced to only 3 (Table \ref{tab:experiments-results-reduced}), the increase in performance changes. 

\begin{figure}[!htp]
    \centering
    \includegraphics[width=1\linewidth]{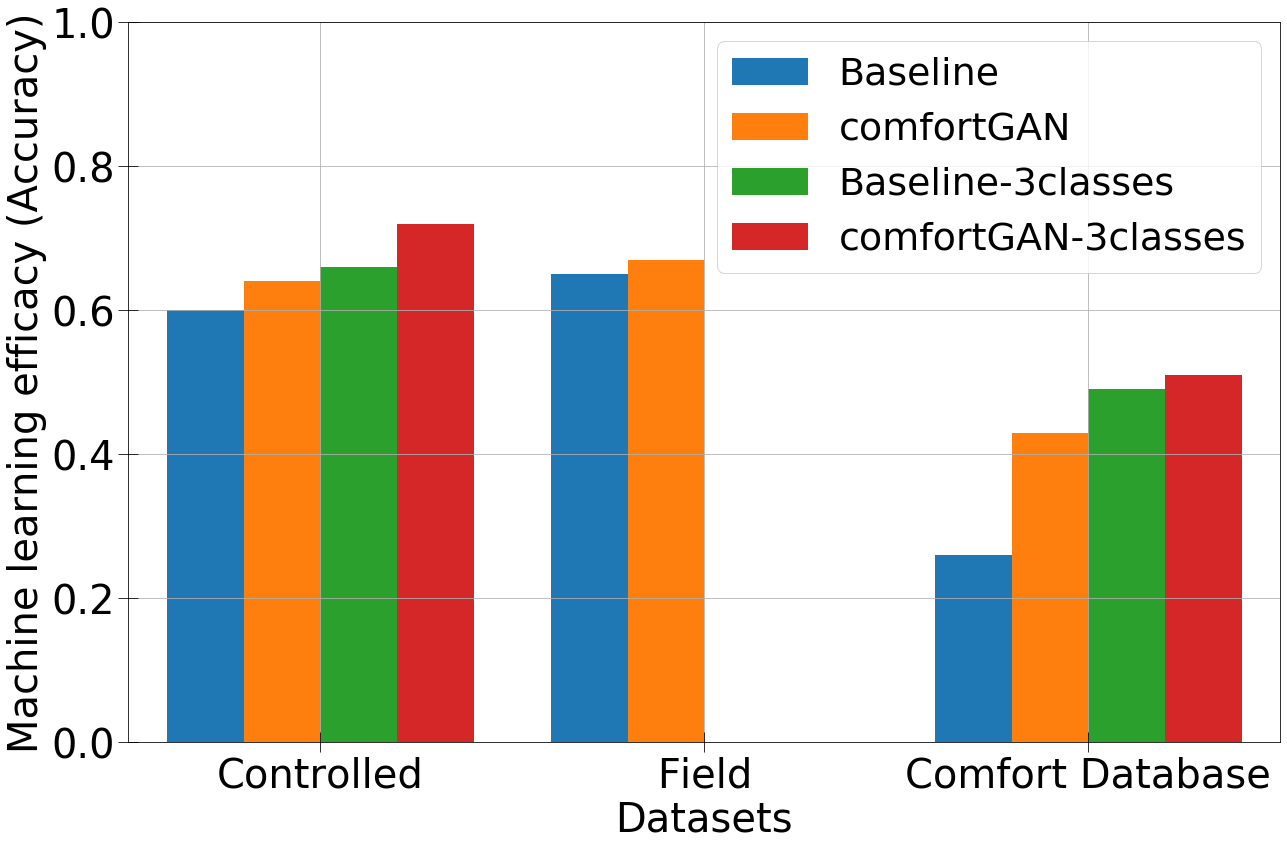}
    \caption{Machine learning efficacy (F1-micro score) of the baseline and comfortGAN approaches for all datasets with their original number of classes and with reduced classes (\textit{-3classes})}
    \label{fig:accuracies}
\end{figure}

For the controlled dataset \cite{Francis2019_dataset}, where classes are reduced from 5 to 3, the increase in performance by comfortGAN is sort of maintained (6\% increase), but for the Comfort Database \cite{FoldvaryLicina2018} where classes are reduced from 7 to 3, comfortGAN performance increase is barely 2\%, while all the other methods provide 1\% increase. The field dataset \cite{Jayathissa2020}, which originally had three classes, to begin with, also shows a minimal increase of 2\% in performance by comfortGAN, while the other GAN variants achieve at most almost the same as the baseline performance. Figure \ref{fig:accuracies} summarises this performance, in terms of F1-micro score, of the baseline and comfortGAN approaches for all datasets with their original and reduced classes. 

\section{Discussion} 
\label{discussions}
Though GAN variants, particularly our proposed method comfortGAN, surpassed the traditionally used augmentation algorithms for thermal comfort model classification performance, they are not ready to be a one-size-fits-all easily implementable solution. CTGAN \cite{CTGAN-Xu2019}, TGAN \cite{Xu2018a}, and our proposed comfortGAN models required some preliminary fine-tuning and longer training time, unlike the traditionally used SMOTE \cite{smote} and ADASYN \cite{adasyn} that are available to building analysts and researchers for immediate use. However, even after going through the process of utilizing these more complex models, the results in Figure \ref{fig:accuracies} show that class balancing for thermal comfort modeling achieves a small performance increase when compared to their baseline.

\subsection{Evaluation and effect on classification performance}
\label{eval_class_discuss}
The assessment of generated samples with Euclidean distance (\textit{Variability} and \textit{Diversity} in Figure \ref{fig:evaluation_diagrams}) may not be as a reliable similarity metric as the SSIM score is for images in the field of computer vision. The GAN related literature usually prefers metrics such as Inception Score or Fr\'echet Inception distance for comparing different GAN variants \cite{Lucic2017}; the former measures the ability of the generative model to retain the class ratio, and the latter measures the distribution of features and labels, but it's biased when there's a small number of samples per class. Based on this, we opted to retain the Euclidean distance-based metric that can be used with the traditionally used methods (SMOTE \cite{smote} and ADASYN \cite{adasyn}).

Moreover, unlike other studies of GANs in the built environment \cite{Yan2020} that constructs a balanced test set for evaluation, we decided to randomly split the datasets into a train and test split, which lead to an also imbalanced test set, in order to mimic what is done in practice by thermal comfort researchers. However, in this scenario, a classification model still gives more importance to the predominant class in the test set and holds back a wider evaluation of the generated under-represented classes. Also, it is possible that a more tuned model on each balanced dataset performs better than the current results in Table \ref{tab:experiments-results}, but we fixed an RDF for each dataset to assess the impact on the generated samples alone. As mentioned in \cite{Theis2015}, a proper assessment of generative model performance is only possible in the context of the application. For the thermal comfort model context, if the goal is to use the model's prediction for building operations control, the GAN performance assessment is tied with the model's prediction performance. However, quantitative evaluation metrics for GANs is currently a critical research direction \cite{Arora2018, Sajjadi2018, odena2019open}.

Additionally, as we show in Figure \ref{fig:accuracies}, when the dataset has only three classes, the classification performance of a balanced dataset, even with the top-performing method (which is based on our results in Table \ref{tab:experiments-results-reduced} is comfortGAN), barely increases. This result raises the question of whether class balancing is needed at all. As mentioned in Section \ref{introduction}, thermal comfort models are ultimately used for better design of indoor spaces or to modify current control strategies in indoor environments \cite{Park2018, Jung2020}. Therefore, from a human perspective, the difference between predicting ``slightly cool'' instead of ``cool'' is smaller than predicting ``slightly cool'' instead of ``neutral''. The baseline performance increase on the controlled dataset \cite{Francis2019_dataset} and the Comfort Database \cite{FoldvaryLicina2018} by reducing the number classes is 6\% and 23\% respectively, surpassing what the augmentation methods could provide (4\% and 17\% respectively). Although using augmentation methods on the datasets with reduced classes can still give more performance improvement, this performance increase seems to be minimal in two datasets, 1\% on the field dataset \cite{Jayathissa2020} and 2\% on Comfort Database \cite{FoldvaryLicina2018}. Even if an increase of 6\% can be achieved on a dataset with reduced classes (comfortGAN on the controlled dataset \cite{Francis2019_dataset}, Table \ref{tab:experiments-results-reduced}), an alternative in a real-world application to improve the performance of a thermal comfort classification model is to reduce the number of classes. By doing this as a first step, we simplify the classification problem and might even be able to ditch the class-balancing pre-processing to focus on other more promising aspects (e.g., better models themselves) to increase classification performance. This idea can be further explored on a control strategy's energy and thermal comfort repercussions with a 5 or 7 class thermal comfort model compared to a control strategy that relies on a 3 class thermal comfort model. We leave this direction as future work.

\subsection{Model applicability}
Considering the limitations and downsides of GANs when deployed in practice, e.g., training stability, hyper-parameter tuning, etc., GANs are still a popular approach for data generation, particularly when dealing with human-generated samples. The Generator does not have access to the real data during the entire training process, making privacy more achievable than other used methods like Variational Autoencoders \cite{pateGAN-Jordon2019}. In the built environment, privacy and anonymization of thermal comfort responses is not necessarily a concern since occupants willingly share this information with their peers or the facilities manager. While the usability of GANs for class-balancing can be circumvented by reducing the number of classes in the dataset, their upper-hand in implicitly learning distributions can be used for other tasks like transfer learning (e.g., learning the representation of a bigger thermal comfort experiment to be applied on a smaller one), self-labeling of samples (e.g., use the data points with subjective feedback to label the measurements without feedback), or missing data imputation (based on learning the data distribution first). We leave the exploration of these options as future work.

\subsection{Class balancing}
Finally, the main assumption made on this balancing scenario is that the data available to us is, to some extend, not fully representative of the reality, and the data is disproportionate due to the collection methods. If the real data is actually balanced in nature, researchers could collect more data or augment the already collected data. However, if this is not the case and the real underlying distribution is, in fact, disproportionate, by augmenting it or re-balancing it, we are changing its nature. In the thermal comfort context, one of the main reasons we obtain imbalanced datasets, in field experiments at least, is because it is expected that occupants find their thermal environment acceptable in most situations. In fact, a balanced dataset from a field experiment would indicate that the building is very poorly managed since it's allowing such a range of situations.

Thus, a modification on the data collection process, mentioned in Section \ref{introduction}, is to conduct targeted surveys to purposely query the user for subjective comfort feedback responses in unseen scenarios. E.g. if not enough `cold` responses are available, a survey will be prompted to the user when the temperature surrounding it is below 15$^{\circ}$C or another arbitrary low temperature. Initial results of this approach in \cite{targeted-surveyDuarte2020} show that this type of surveying method decreases user disturbance and data redundancy. Furthermore, relying on peer information can also alleviate the need to collect longitudinal data from every single building occupant \cite{Jayathissa2020}.

Finally, class balancing is an alternative to address imbalance datasets. As mentioned in Section \ref{eval_class_discuss}, in the thermal comfort context, the difference between the class values (e.g., ``slightly cool'', ``cool'', and ``warm'') might be better captured as a cost-sensitive classification problem or reformulation on how these labels are seen by the loss functions of a given model. These other approaches, however, are out of the scope of this work.

\section{Conclusions}
\label{conclusions}
In this work, we present comfortGAN, a conditional Wasserstein GAN-based approach for data augmentation, specifically class-balancing, in thermal comfort datasets. This approach's main contribution is the appropriate combination of GAN architecture and loss function combined with an extensive evaluation on three publicly available thermal comfort datasets. The datasets cover controlled and field experiments as well as the Comfort Database (the biggest thermal comfort database available to date). We evaluated the effect on performance a balanced dataset, comprised of generated and real samples, has on the thermal comfort classification task. Mimicking building analysts and thermal comfort researchers modeling set up, we evaluated samples generated from comfortGAN, two traditionally use methods (SMOTE \cite{smote} and ADASYN \cite{adasyn}) and two other GAN variants for data synthesis (CTGAN \cite{CTGAN-Xu2019} and TGAN \cite{Xu2018a}). 

We found that GAN variants consistently outperform the traditionally used methods (Table \ref{tab:experiments-results}), and comfortGAN achieves the highest increase in performance: from 4\% in the controlled dataset \cite{Francis2019_dataset} to 17\% in the Comfort Database \cite{FoldvaryLicina2018}. However, reducing the number of classes in the dataset by merging similar values (for example, ``cold``, ``cool'', and ``slightly cool'') already increases the baseline classification performance (6\% in the controlled dataset \cite{Francis2019_dataset} and 23\% in the Comfort Database \cite{FoldvaryLicina2018}), reducing the performance increase of augmentation methods to 1\% or 2\% (Table \ref{tab:experiments-results-reduced}). Ultimately, the choice of using a class-balancing algorithm depends on the end goal of the thermal comfort research. For control purposes, a 3 class dataset and model should provide enough information to control building and HVAC systems. Therefore, reducing the number of classes should be the first attempt to have more robust and better classification models, which in some cases might benefit from class balancing (6\% increase in the controlled dataset, Table \ref{tab:experiments-results-reduced}).

Our results open the door for further research into generative models and thermal comfort data. Given that the environmental and physiological parameters are measured at a much higher frequency than the subjective comfort feedback, self-labeling of these data points could be explored with GAN-based approaches or other generative methods as well. Transfer learning on thermal comfort datasets is also a promising venue that could allow researchers to leverage their human studies' cohort size or the number of responses per participant. 

We described comfortGAN as a tool that shows the potential to be part of the data-driven thermal comfort modeling, particularly when compared to the traditional methods commonly used. Even if other practical ways, such as reducing the number of classes, achieve an equal or better performance increase in most cases, we believe the decision relies on the researcher or practitioner and their use case. If their assumptions on the data allow for class-balancing, augmentation methods like comfortGAN can be used as part of their modeling pipeline. On the other hand, if the resulting model will be incorporated into a control strategy, a reduction in the number of classes should be considered first before diving into augmentation methods.



\begin{acks}
This research was funded by the Republic of Singapore's National Research Foundation through a grant to the Berkeley Education Alliance for Research in Singapore (BEARS) for the Singapore-Berkeley Building Efficiency and Sustainability in the Tropics (SinBerBEST) Program. BEARS has been established by the University of California, Berkeley as a center for intellectual excellence in research and education in Singapore
\end{acks}

\bibliographystyle{ACM-Reference-Format}
\bibliography{comfortGAN}

\end{document}